\documentclass[10pt,final,journal]{IEEEtran} 
\usepackage{multirow}  
\usepackage[table]{xcolor}  
\usepackage{arydshln}       
\usepackage{makecell}       
\usepackage[caption=false,font=normalsize,labelfont=sf,textfont=sf]{subfig}
\usepackage{amsmath,amsfonts}
\usepackage{algorithmic}
\usepackage{algorithm}
\usepackage{array}
\usepackage[caption=false,font=normalsize,labelfont=sf,textfont=sf]{subfig}
\usepackage{textcomp}
\usepackage{stfloats}
\usepackage{url}
\usepackage{verbatim}
\usepackage{graphicx}
\usepackage{cite}
\usepackage[hidelinks]{hyperref}

\makeatletter
\def\@IEEEtablecaption#1#2{%
  \@makecaption{#1}{\MakeUppercase{#2}}%
}
\makeatother

\begin{document}

\title{Adversarial Patch Generation for Visual-Infrared Dense Prediction Tasks via Joint Position-Color Optimization}

\author{He Li, Wenyue He, Weihang Kong, Xingchen Zhang$^*$,\IEEEmembership{~Member, IEEE}

\thanks{This paper was supported partly by the National Natural Science Foundation of China (No. 62306264, 62173290), the Central Government Guided Local Funds for Science and Technology Development of China (No. 236Z0303G), the Natural Science Foundation of Hebei Province in China (No. F2024203091, F2025203045), Innovation  Capability Improvement Plan Project of Hebei Province, China (No. 22567626H), the Royal Society Research Grant (No. RG\textbackslash{}R1\textbackslash{}251462), the Royal Society-NSFC International Exchanges Grant (No. W2521174) and Yanshan University (No. 2025LGGH003).

Weihang Kong, Wenyue He, and He Li are with the School of Artificial Intelligence (School of Software), Yanshan University, Qinhuangdao 066004, China. (Email: whkong@ysu.edu.cn; hewenyue@stumail.ysu.edu.cn; lihe@ysu.edu.cn)

Xingchen Zhang is with the Fusion Intelligence Laboratory, Department of Computer Science, University of Exeter, EX4 4RN, United Kingdom. (Email: x.zhang12@exeter.ac.uk)
\newline	$^*$ Corresponding author: Xingchen Zhang }
}

\markboth{Journal of \LaTeX\ Class Files}%
{Shell \MakeLowercase{\textit{et al.}}: A Sample Article Using IEEEtran.cls for IEEE Journals}


\maketitle

\begin{abstract}
Multimodal adversarial attacks for dense prediction remain largely underexplored. In particular, visual–infrared (VI) perception systems introduce unique challenges due to heterogeneous spectral characteristics and modality-specific intensity distributions. Existing adversarial patch methods are primarily designed for single-modal inputs and fail to account for cross-spectral inconsistencies, leading to reduced attack effectiveness and poor stealthiness when applied to VI dense prediction models. To address these challenges, we propose a joint position–color optimization framework (AP-PCO) for generating adversarial patches in visual–infrared settings. The proposed method optimizes patch placement and color composition simultaneously using a fitness function derived from model outputs, enabling a single patch to perturb both visible and infrared modalities. To further bridge spectral discrepancies, we introduce a cross-modal color adaptation strategy that constrains patch appearance according to infrared grayscale characteristics while maintaining strong perturbations in the visible domain, thereby reducing cross-spectral saliency. The optimization procedure operates without requiring internal model information, supporting flexible black-box attacks. Extensive experiments on visual–infrared dense prediction tasks demonstrate that the proposed AP-PCO achieves consistently strong attack performance across multiple architectures, providing a practical benchmark for robustness evaluation in VI perception systems.
\end{abstract}

\begin{IEEEkeywords}
Adversarial patches, Multimodal attacks, Joint optimization, Visible-infrared fusion
\end{IEEEkeywords}

\section{Introduction}
\IEEEPARstart{D}{eep} learning models have achieved remarkable success in a wide range of visual perception tasks and image processing\cite{11369865,11367376,zhang2023visible}. However, despite their strong performance, these models remain vulnerable to adversarial perturbations, making security a growing concern in computer vision \cite{phan2024evaluating,luo2025progressive}. This motivates systematic investigation of the vulnerability of modern perception systems under adversarial perturbations, especially in safety-critical settings.

Visual-infrared (VI) perception has become increasingly important because the two modalities provide complementary cues: visible images capture rich texture and color, while infrared images remain informative under low illumination and adverse weather. 
This complementarity has motivated broad adoption of VI data in dense prediction tasks, such as crowd counting \cite{kong2024multi}, semantic segmentation \cite{chen2025cross}, and image fusion \cite{chen2025sdsfusion}, under diverse environmental conditions.

\begin{figure}
    \centering
    \includegraphics[width=\linewidth]{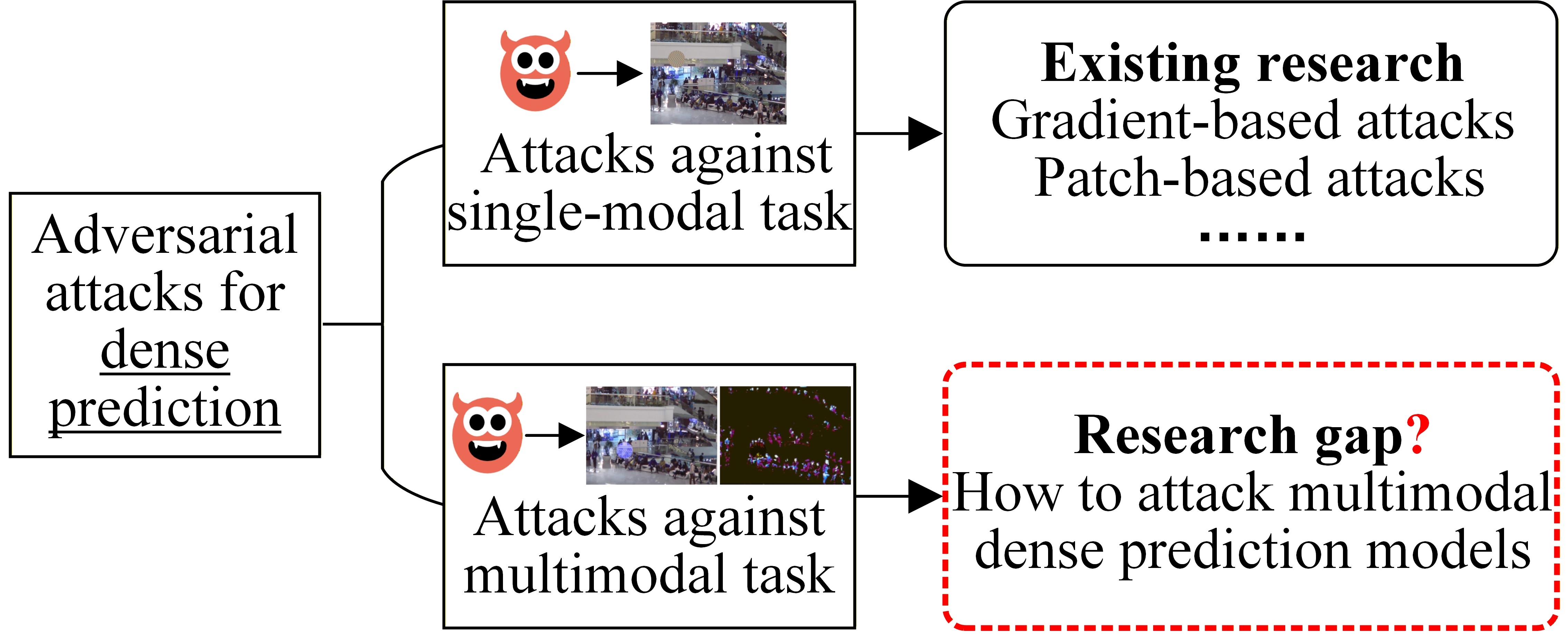}
    \caption{Comparison of existing adversarial attack work and this study. The security of visual–infrared dense prediction models, 
    remains underexplored compared with single-modal settings. This study aims to fill this gap.}
    \label{fig:gap}
\end{figure}

Despite this growing use, the security and robustness of \emph{VI dense prediction} models remain far less studied than their single-modal counterparts, leaving an important yet underexplored dimension of multimodal perception security. 
As illustrated in Fig.~\ref{fig:gap}, existing adversarial research has predominantly concentrated on single-modal settings, with limited attention paid to VI dense prediction scenarios.

\begin{figure*}
	\centering
	\includegraphics[width=1\textwidth]{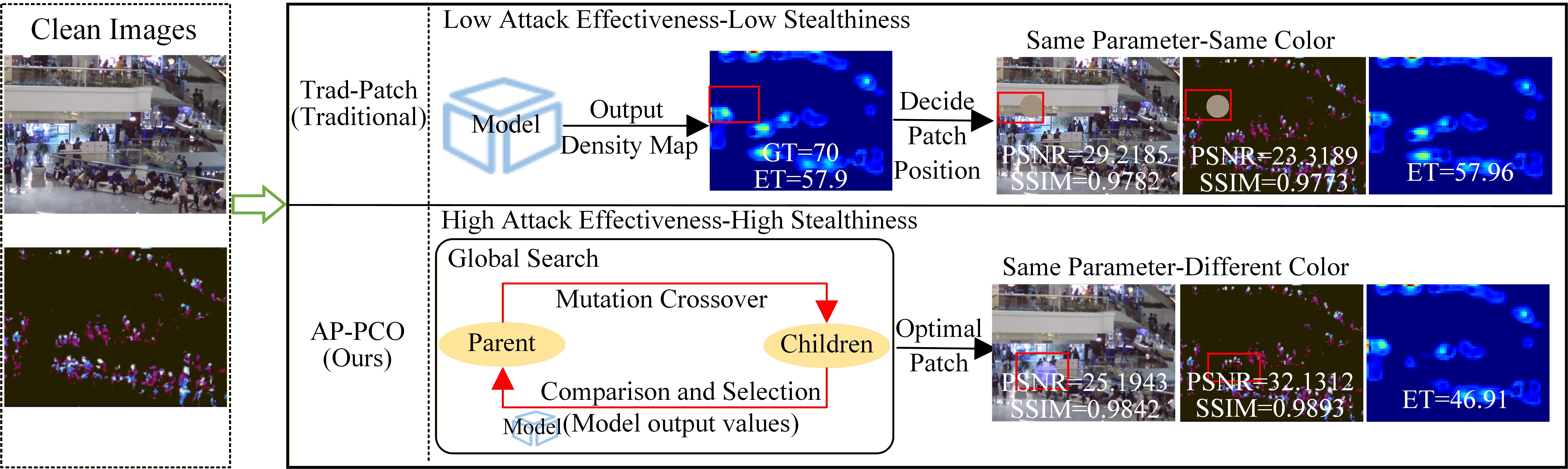}
     \caption{Comparison between traditional adversarial patches (“Trad-Patch”) and the proposed patch (“AP-PCO”). 
    Traditional methods determine the patch position based on a single forward pass of the target model and do not optimize color for multimodal data, which limits their effectiveness and stealthiness in visual–infrared settings. In contrast, our approach performs iterative, gradient-free global search to jointly determine position and color, and incorporates a cross-modal color reuse strategy to achieve stronger attacks and better stealthiness.}
    \label{int}
\end{figure*}

 As a consequence, most existing adversarial attacks are developed under single-modal assumptions, primarily focusing on visible images \cite{li2021generative, liu2022harnessing}. 
When directly applied to VI dense prediction tasks, these methods often exhibit degraded effectiveness due to substantial spectral differences between visible and infrared inputs. Dense prediction further increases the difficulty of attack design because model responses are spatially distributed over dense outputs.Consequently, adversarial patterns effective in the visible domain may not reliably transfer to VI dense prediction models.
Among various attack forms, adversarial patches are especially relevant because they are physically realizable and offer a flexible carrier for modality-aware perturbation design \cite{shack2024breaking}. Their effectiveness mainly depends on two factors: spatial deployment (position and coverage) and spectral appearance design (e.g., color/texture). Jointly optimizing these two factors is critical for achieving strong yet stealthy attacks, especially in VI dense prediction where spatial placement and appearance jointly influence pixel-level responses. Although several studies attempt joint optimization \cite{ma2023transferable, wu2024uniid}, most of them still treat spatial and content updates in loosely coupled or sequential manners, leading to limited generalization in VI dense prediction settings.

For position optimization, most existing works rely heavily on task-specific information (e.g., bounding boxes in object detection or density peaks in crowd counting), which does not generalize to dense prediction models without explicit region proposals. This lack of transferable cues makes it difficult to locate effective attack regions, especially in heterogeneous visual–infrared settings, as shown in Fig.~\ref{int}.

Regarding content optimization, many existing methods rely on GAN-based iterative strategies \cite{chakraborty2024ten}. These methods need large-scale training data to learn content distributions and generation mappings, entailing complex training. Existing techniques mainly include texture-based \cite{yang2020design, ma2023transferable} and color-based optimization \cite{wu2024uniid}. Texture methods can generate complex patterns but become inefficient and unstable in visual–infrared settings because of the fundamental differences between the modalities. Color optimization is more efficient, as it avoids modeling fine-grained structures, but current color-mapping designs do not adapt across modalities: patches optimized for visible images often fail to transfer to infrared images and produce noticeable artifacts (Fig.~\ref{int}), making it difficult to balance attack strength and stealthiness.

Furthermore, although some methods attempt joint optimization, spatial and appearance updates are typically implemented in sequential or loosely coupled manners, limiting effective coordination between the two factors. 
Such weak coupling is particularly problematic for visual–infrared dense prediction, where spatial placement and appearance jointly influence pixel-level model responses.

In addition, adversarial patches are parameterized by a mixed discrete–continuous structure: spatial parameters determine a binary mask, while color parameters define continuous perturbations within the selected region. This mixed parameterization further complicates the optimization landscape in cross-modal settings, making effective joint perturbation design inherently challenging.

To address these challenges, we propose a global search mechanism that jointly optimizes patch position and color via simulating population evolution in a high-dimensional solution space. This mechanism reduces reliance on task and model-specific cues and achieves better generalization across visual–infrared dense prediction tasks by avoiding separate mismatched optimization. Additionally, we present a cross-modal color reuse strategy (mask multiplication, grayscale compression, background superposition) to resolve infrared image artifacts (Fig.~\ref{int}). It ensures cross-modal color consistency, enables the patch to naturally integrate with infrared grayscale characteristics while maintaining high brightness in visible images, and balances attack strength with stealthiness.

The main contributions of this paper are as follows:
\begin{itemize}
\item We formulate VI dense prediction patch attacks as a joint spatial-spectral optimization problem and propose a population-based global search mechanism to jointly optimize patch position and color for visible and infrared inputs, achieving consistent attack performance across different models and tasks.

\item We introduce a cross-modal color parameter reuse strategy that adapts a shared appearance representation to both visible and infrared modalities, effectively reducing infrared saliency while maintaining strong perturbations in the visible domain.

\item We conduct comprehensive experiments on three representative VI dense prediction tasks (i.e., crowd counting, semantic segmentation, and image fusion) and evaluate the proposed method under multiple defense strategies. The results demonstrate consistent attack effectiveness across diverse architectures, tasks, and defense settings.
\end{itemize}

\section{Related work} 
\label{related_work}

\subsection{Visual-infrared perception}
Visible and infrared images have been widely applied in perception tasks such as crowd counting \cite{kong2024multi}, semantic segmentation \cite{chen2025cross}, and image fusion \cite{chen2025sdsfusion}. Their complementary sensing properties, where visible images capture detailed textures and infrared images provide stable thermal information under low illumination, make visual–infrared perception systems essential for all-weather computer vision. However, the multimodal nature of these systems also presents unique challenges for AI security because perturbations may affect visible and infrared representations in different ways.

Despite the growing importance of visual–infrared perception, very few studies have examined adversarial attacks in this setting. The robustness of multimodal perception models therefore remains insufficiently understood, leaving a clear gap for systematic investigation.

\subsection{Adversarial attacks}

Adversarial attacks aim to mislead a trained model during inference by introducing carefully crafted perturbations to the input data \cite{abomakhelb2025comprehensive}. Depending on the attacker’s knowledge of the target model, attacks are typically categorized into white-box and black-box settings. White-box attacks \cite{guo2024regressor} assume full access to model parameters and gradients, while black-box attacks \cite{cheng2024feature} rely on query-based optimization or transferability without requiring internal model information.

From the implementation perspective, attacks can occur in the digital domain \cite{wu2024adversarial, barik2024adversarial} or the physical domain \cite{wei2024physical, fang2024unified}. Adversarial patches form an important bridge between the two, as they can be optimized digitally and deployed physically with high practicality. Brown et al.~\cite{brown2017adversarial} first introduced adversarial patches for single-modal image recognition, leading to a series of patch-based methods for visible images, such as APAM \cite{wu2021towards} and PAP \cite{liu2022harnessing}. Sun et al.~\cite{sun2024adversarial} further demonstrated that local patch manipulation can significantly degrade image fusion models. However, all these approaches operate exclusively in the visible modality.

Research on adversarial attacks for VI dense prediction remains extremely limited. The perception fusion framework \cite{liu2023paif} employs PGD to generate adversarial samples, but relies on gradients and does not support patch-based physical deployment. Motivated by this gap, this work focuses on visual–infrared cross-modal tasks and aims to design adversarial patches affecting both modalities simultaneously.

\subsection{Positions and content optimization of patches}
Given the significant influence of patch position and content on attack performance, many studies have investigated their roles in adversarial patch design. For example, He et al. \cite{he2024dorpatch} distributed multiple adversarial patches across images, where position optimization was performed jointly for all patches, and structural loss was introduced to guide content optimization. Yang et al. \cite{yang2020design} optimized the pixel values of adversarial patches through gradient ascent. Wu et al. \cite{wu2024uniid} optimized patch positions through key facial regions and refined patch content using gradient-based algorithms with multi-objective loss functions. Li et al. \cite{li2021generative} employed a generator–decoder architecture to produce constrained and dynamically transformed patch positions and textures, achieving end-to-end optimization through weighted fusion. Ma et al. \cite{ma2023transferable} proposed a spatially mutable adversarial patch method that identifies key positions with the greatest impact on target identity using gradient search and iteratively optimizes patch textures with a texture gradient loss.

Despite their progress, these methods rely heavily on complex generators or internal model information and are constrained by the feature representation of specific detection models and predefined key regions, making it difficult to achieve flexible adaptation in complex scenarios. 
\begin{figure*}
	\centering
	\includegraphics[width=1\textwidth]{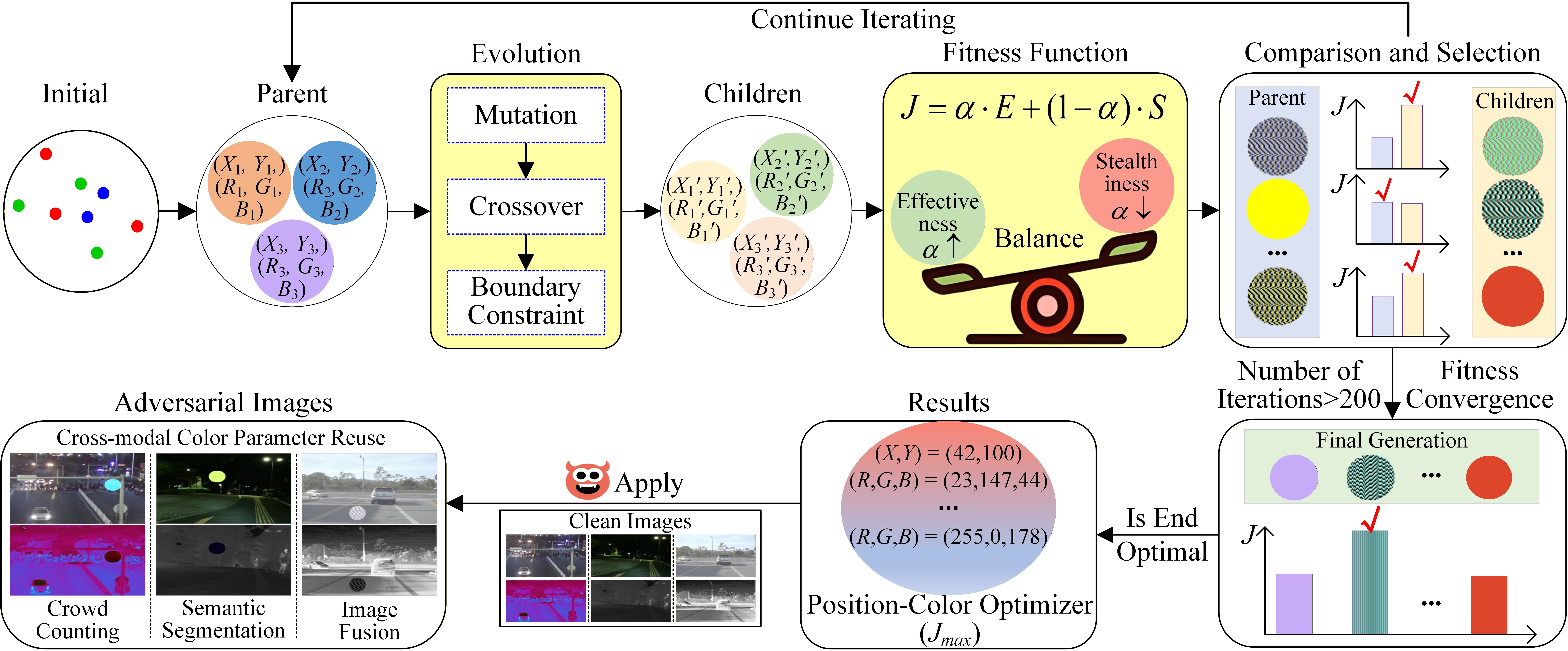}
	\caption{Framework of cross-modal adversarial patches with position-color joint optimization. The initial population consists of a series of randomly generated circular samples. Then, through mutation, crossover, and boundary handling, a child population with diverse positions and colors is generated. A fitness function is then applied for cross-modal evaluation to compare the parent and child populations and repeat the iterative loop, ultimately obtaining the optimal patch deployed for dense prediction tasks.}
	\label{zhutu}
\end{figure*}

\section{Methodology} 
\label{Methodology}

\subsection{Threat model}
This paper conducts research on adversarial attacks against VI 
dense prediction tasks, with the core objective of generating adversarial patches with excellent attack effectiveness and relative stealthiness to interfere with the model outputs of three representative tasks, i.e., crowd counting, semantic segmentation, and image fusion. A black-box attack assumption is adopted in this research, the attacker does not need to acquire complete information such as the internal architecture and parameter configuration of the target model, and can implement effective attacks merely through output feedback. 

Based on the above threat model, the paper proposes a global optimization framework that jointly searches for adversarial patches optimal position and color in a unified parameter space, guided by a fitness function and optimized via a global search mechanism. Its pipeline: encode patches into parameter vectors (position and color), generate initial candidate populations, iteratively update via global search (evaluate candidates using the fitness function, retain higher-fitness ones), and decode the best vector for the final patch upon convergence or max iterations. For clarity, the overall framework is illustrated in Fig \ref{zhutu} and the patch generation procedure is detailed in Algorithm \ref{algorithm}.

\begin{algorithm}
\begin{small}
	\caption{Generate Adversarial Patch} 
    \renewcommand{\algorithmicrequire}{\textbf{Input:}}
    \renewcommand{\algorithmicensure}{\textbf{Output:}}
	\label{algorithm} 
	\begin{algorithmic}[1]
	\REQUIRE Clean visible image $X_{vis}$, clean infrared image $X_{inf}$, the fitness function $J(\cdot)$, the max number of iterations $T$
	\ENSURE Visible adversarial example $X_{vis}^{adv}$ and infrared adversarial example $X_{inf}^{adv}$;
	\STATE Initialize Population $S(0)$
	\STATE \quad \textbf{for} $k = 0$ to $T$-1 \textbf{do}  
	\STATE \qquad Sort $S(k)$ in descending order according to $J(S(k))$
	\STATE \qquad \textbf{if} $S_0(k)$ makes the attack successful \textbf{then} 
	\STATE \qquad \quad stop = $k$; break;
	\STATE \qquad \textbf{end if}
	\STATE \qquad Generate $S(k + 1)$ based on crossover and mutation. 
	\STATE \qquad Limit boundaries of $S(k + 1)$
	\STATE \qquad \textbf{for} $i = 1$ to $Q$ \textbf{do} 
	\STATE \qquad \quad Evaluate $S_i(k)$ and $S_i(k+1)$  
	\STATE \qquad \quad $S_i(k+1) \leftarrow$ the better one in $S_i(k)$ and $S_i(k+1)$
	\STATE \qquad \textbf{end for} 
	\STATE \quad \textbf{end for}
	\STATE Sort $S(stop)$ in descending order according to $J(S(k + 1))$
	\STATE Choose $S_0(stop)$ as the final individual from $S(stop)$
	\STATE Generate unified patch $M$ with $S_0(stop)$ by integrating position $(x,y)$ and color $(R,G,B)$
	\STATE Obtain adversarial examples with $M$ 
	\RETURN $X_{vis}^{adv}$, $X_{inf}^{adv}$
	\end{algorithmic} 
    \label{algorithm}
    \end{small}
\end{algorithm}

\subsection{Problem formulation}
In visual–infrared dense prediction tasks, a joint cross-modal adversarial attack takes a pair of clean visible and infrared images as input and generates perturbed counterparts by embedding an adversarial patch. The objective of the adversarial patch is to perturb the input in a way that interferes with the model output.

Let $X_{vis}$ and $X_{inf}$ denote the original visible and infrared images of size $h \times w$. The adversarial images $X_{vis}^{adv}$ and $X_{inf}^{adv}$ are defined as
\begin{equation}
X_{vis}^{adv} = X_{vis} \odot (1-M) + X_{vis}' \odot M,
\label{1}
\end{equation}
\begin{equation}
X_{inf}^{adv} = X_{inf} \odot (1-M) + X_{inf}' \odot M,
\label{2}
\end{equation}
where $M \in \{0,1\}^{h \times w}$ is a binary mask indicating the patch region. A value of $M_{ij}$=1 means that pixel $(i,j)$ belongs to the patch, while $M_{ij}$=0 preserves the original image. The matrices $X_{vis}'$ and $X_{inf}'$ specify the pixel values of the patch in the visible and infrared modalities, respectively, and $\odot$ denotes the Hadamard product.

\subsection{Global Optimization Framework}
In cross-modal patch optimization, the underlying search space exhibits highly non-convex and partially discontinuous characteristics. 
The spatial parameters $(x,y,r)$ determine a binary mask $M$, which introduces discrete structural changes in the input space, while the color parameters define continuous perturbations within the selected region. 
This mixed discrete-continuous parameterization leads to a rugged fitness landscape with multiple local optima.

Furthermore, visual and infrared modalities possess heterogeneous spectral responses. 
A perturbation that induces strong activation changes in the visible branch may produce weak or inconsistent responses in the infrared branch. 
Such modality-dependent sensitivity further increases the irregularity of the optimization landscape. 
In dense prediction models, outputs are spatially aggregated across feature maps. 
Small changes in patch placement may therefore activate different local receptive fields, causing abrupt variations in the fitness landscape. 
As a result, the optimization problem becomes highly non-linear and sensitive to spatial placement.

Under the adopted black-box threat model, internal gradients and model parameters are unavailable. 
Even if numerical gradient approximation is applied, the discontinuity introduced by binary masks and modality-specific feature extraction may result in unstable or misleading update directions. 
Therefore, local gradient-based optimization methods are not well suited for exploring the combined spatial--spectral parameter space.

To address these challenges, we adopt a population-based global optimization strategy that can effectively explore disjoint feasible regions and escape local optima. 
Population-based methods maintain multiple candidate solutions simultaneously and update them through stochastic variation, enabling broader exploration of the search space.

Specifically, Differential Evolution (DE) is employed as the optimization backbone due to its strong global search capability, robustness to irregular objective landscapes, and suitability for mixed-parameter optimization. 
Within this framework, each individual encodes a complete set of patch parameters, including spatial position and color values. 
The population is iteratively refined according to the fitness function described in Section III-D, and the candidate with the highest fitness is selected as the final adversarial patch.

DE refines a population of candidate solutions through mutation, crossover, and selection. 
Compared with traditional evolutionary algorithms such as genetic algorithms or particle swarm optimization, DE provides faster convergence and stronger global exploration ability in high-dimensional parameter spaces. 
These properties make DE particularly suitable for optimizing adversarial patches in VI 
 dense prediction tasks.

\subsection{Fitness function}
\label{subsec:fitness-function}
The optimization process is guided by a fitness function that evaluates the progress of cross-modal adversarial attacks. By measuring both the attack strength of the patch and its stealthiness, the fitness function provides a clear direction for the iterative evolution of patch parameters, defined as
\begin{equation}
J = \alpha \cdot E(X^{adv}) + (1 - \alpha) \cdot S(X^{adv}),
\label{18}
\end{equation}
where $\alpha \in [0,1]$ controls the trade-off between attack effectiveness and stealthiness. A larger value of $\alpha$ places more weight on the effectiveness term $E(\cdot)$ and encourages the optimization to produce stronger attacks, while a smaller value places more emphasis on the stealthiness term $S(\cdot)$ and encourages less noticeable patches.  The choice of $\alpha$ may be set through validation or adjusted according to specific attack–defense requirements. $E$ denotes attack effectiveness, which has different formulations in various dense prediction tasks. 
 
 In the crowd counting task, $E$ refers to the people count error (as shown in Eq. \ref{19}):
 \begin{equation}
 \begin{aligned} 
{{E}_{count}}=|count(X^{adv})-count({{X}})|,
 \end{aligned}
 \label{19}
\end{equation}
where $X$ denotes the original image, $X^{adv}$ denotes the adversarial image, $count(\cdot)$ denotes the function for calculating the number of people.
 
 In the semantic segmentation task, it refers to the mIoU (as shown in Eq. \ref{20}):
 \begin{equation}
 \begin{aligned} 
{{E}_{seg}}=100-(mIoU(X^{adv})\times 100),
 \end{aligned}
 \label{20}
\end{equation}
where $mIoU(\cdot)$ denotes the function for calculating the mIoU value.

In the image fusion task, it uses gradient loss ($\mathcal{L}\_grad$), intensity loss ($\mathcal{L}\_inten$), and SSIM, as shown in Eq. \ref{22}:
\begin{equation}
 \begin{split} 
{{E}_{fus}}&=\mathcal{L}\_inten({{X}_{vis}},{{X}_{\inf }},X^{adv})\times 20\\
&+\mathcal{L}\_grad({{X}_{vis}},{{X}_{\inf }},X^{adv})\times 20\\
&+(1-SSIM({{X}_{vis}},{{X}_{\inf }},X^{adv}))\times 10,
\end{split}
 \label{22}
\end{equation}
where  ${X}_{vis}$ and ${X}_{inf}$ denote the original visible image and the infrared image, respectively. 

 $S$ denotes stealthiness, employing SSIM and PSNR (Eq. \ref{21}) \cite{hore2010image} in the crowd counting and semantic segmentation tasks. 
\begin{equation}
\begin{split}     
{{S}_{count}}/{{S}_{seg}}&=({PSNR_{vis}}+{PSNR_{\inf }})\times 1\\
&+({SSIM_{vis}}+{SSIM_{\inf }})\times 20,
 \end{split}
 \label{21}
\end{equation}
where $PSNR_{vis}$ represents the PSNR value between the original image and the adversarial image in the visible modality, $SSIM_{vis}$ represents the SSIM value between the original image and the adversarial image in the visible modality, $PSNR_{inf}$ represents the PSNR value between the original image and the adversarial image in the infrared modality, and $SSIM_{inf}$ represents the SSIM value between the original image and the adversarial image in the infrared modality.

For crowd counting and semantic segmentation, attack effectiveness is evaluated using task-specific metrics (e.g., GAME, RMSE, mIoU), while stealthiness is measured with SSIM and PSNR, as these two aspects are negatively correlated.
However, for the image fusion task, SSIM and PSNR already serve as the indicators of task performance, since fusion effectiveness inherently depends on the similarity between the fused image and the original inputs.
Therefore, using SSIM or PSNR again to quantify stealthiness would create a coupling between attack effectiveness and stealthiness, and thus the stealthiness metric is not considered for the fusion task.

\subsection{Representation of patch position }
The position of the adversarial patch is parameterized by a unified center coordinate $(x, y)$ and a radius $r$, which determines the patch size. To ensure that the patch remains within image boundaries, the feasible region of the patch center is constrained by
\begin{equation}
x \in [\,r,\; w - r\,], \quad y \in [\,r,\; h - r\,],
\label{14}
\end{equation}
where $w$ and $h$ denote the width and height of the image, respectively.

\begin{figure*}
	\centering
	\includegraphics[width=1\textwidth]{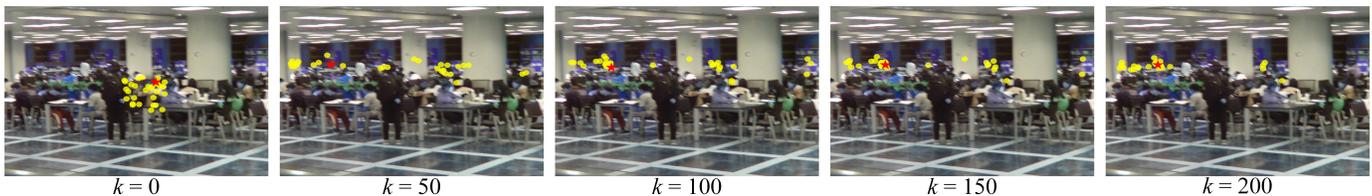}
	\caption{Variation trajectory of patch positions across iterations. Yellow markers denote the patch center coordinates of all individuals in each generation, while red markers indicate the center coordinate of the individual with the highest fitness in that generation. As iterations progress, the population gradually converges toward crowd-dense regions and ultimately stabilizes at the optimal patch location.}
	\label{diedai}
\end{figure*}    

\subsection{Representation of patch color}
In addition to position, each adversarial patch is assigned a list of color values, where each color is represented by an $[R,G,B]$ triplet. After localizing all pixels within the circular patch region, the colors in the list are assigned sequentially to each pixel using a modulo operation. This cyclic assignment produces a multi-color pattern without increasing the dimensionality of the parameter space.

Visible and infrared images differ substantially in appearance and dynamic
range, so directly applying visible colors to infrared images often
produces unrealistic artifacts. To address this problem, we adopt a
cross-modal color parameter reuse strategy. In the visible modality, the color mask is multiplied with the three-channel image to form a high-brightness
region that strongly disrupts texture and color cues. In the infrared
modality, the same color parameters are first converted to grayscale and then
compressed in intensity, allowing the patch to blend naturally into the
grayscale appearance of infrared images and avoiding abrupt visual patterns.
This strategy improves stealthiness while maintaining strong
perturbation capability across modalities.

\subsection{Joint optimization}

The optimization of patch position and color is performed jointly within a unified parameter vector. For each individual in the DE population, the vector contains both spatial parameters $(x, y, r)$ and color parameters $([R,G,B]$ values). The initial parent population is constructed from randomly generated patches. During optimization, for each $V_m$ in the $k$-th population, three distinct individuals $V_{m1}$, $V_{m2}$, and $V_{m3}$ are randomly selected to generate a mutation vector
\begin{equation}
V_{mut} = V_{m1} + f \cdot ( V_{m2} - V_{m3} ),
\label{15}
\end{equation}
where $f$ is the scaling factor controlling the influence of the differential pair. The mutation vector is combined with $V_m$ through binomial crossover to form a trial vector
\begin{equation}
V_{trial}(j) =
\begin{cases}
V_{mut}(j) & \text{if } \text{rand}(0,1) \le CR \text{ or } j = J_{rand} \\
V_m(j) & \text{otherwise}
\end{cases}
\label{16}
\end{equation}
where $CR \in [0,1]$ is the crossover probability, $j$ denotes a parameter dimension, and $J_{rand} \in \{1,2,\ldots, D\}$ is a randomly selected index ensuring at least one mutated dimension. This process allows the search space to be explored effectively while preserving useful characteristics of the current solution.  The fitness function defined in Section~\ref{subsec:fitness-function} is then computed for both the $V_{trial}$ and the $V_m$, and the one with higher fitness is retained. 
This iterative process continues until the population converges or a fixed
number of iterations (e.g., 200) is reached. As shown in Fig.~\ref{diedai}, the individuals gradually converge toward the most effective patch location, and
the color parameters are updated according to the balance between attack
effectiveness and stealthiness defined by the weighting parameter $\alpha$.
The final output of the algorithm is the adversarial patch with the highest
fitness value across the visible and infrared modalities.

\subsection{Optimization Coupling Analysis}

The patch generation process described previously can be interpreted as solving a joint optimization problem with respect to both spatial and color parameters. 
Specifically, guided by the fitness function defined in Eq.~(3), the optimization objective can be written as
\begin{equation}
\max_{x,y,r,\mathbf{c}} 
J\big(X^{adv}_{vis}, X^{adv}_{inf}\big),
\end{equation}
where $(x,y,r)$ denote the spatial parameters determining the patch mask $M$, and $\mathbf{c}$ denotes the set of color parameters controlling the patch content in both modalities.

As defined in Eqs.~(1)--(2), the adversarial images are constructed by embedding the patch mask $M(x,y,r)$ and the corresponding color-dependent patch contents into the original visible and infrared images. 
Therefore, the fitness value $J$ is an implicit function of both the spatial support region and the color-induced perturbation magnitude.

In dense prediction tasks, model outputs are computed from spatially distributed feature responses. 
The perturbation introduced by an adversarial patch affects a set of pixels determined by the mask $M(x,y,r)$, and its influence on the fitness function can be approximated as
\begin{equation}
\Delta J \approx 
\sum_{(i,j)\in M(x,y,r)}
\frac{\partial J}{\partial X_{i,j}} \cdot \delta X_{i,j},
\end{equation}
where $\delta X_{i,j}$ is determined by the color parameters $\mathbf{c}$.

This expression reveals the intrinsic coupling between spatial and color parameters. 
The spatial parameters $(x,y,r)$ determine which pixel locations contribute to the perturbation, while the color parameters determine the direction and magnitude of perturbation at those locations. 
Consequently, the effect of a given color configuration depends on the selected spatial region, and vice versa.

Optimizing position while fixing color restricts the perturbation direction in feature space, whereas optimizing color under a fixed position limits the exploitable spatial sensitivity of the model. 
Independent optimization therefore constrains the reachable perturbation subspace and may lead to suboptimal local solutions. 
This observation implies that the optimization problem is inherently inseparable in spatial and spectral dimensions.

This coupling effect becomes more pronounced in VI settings. 
The visible modality operates in a three-channel RGB space, whereas the infrared modality primarily relies on intensity information. 
Furthermore, multimodal fusion modules aggregate spatial and spectral cues jointly. 
As a result, the sensitivity of $J$ to perturbations depends simultaneously on spatial placement and cross-spectral consistency. 
Joint optimization enables coordinated exploration in the combined spatial--spectral parameter space, leading to stronger cross-modal perturbations and improved attack effectiveness in dense prediction tasks.

\section{Experiments} \label{exp}
\subsection{Experimental settings}
\subsubsection{Implementation details}

During the iterative optimization process, the fitness function was set to consider only attack effectiveness (\(\alpha = 1\)) for the main experiments. This unified setting allows fair comparison across models and tasks, as the primary goal of these experiments is to evaluate the attack effectiveness of the proposed method under a consistent configuration. To demonstrate the flexibility of the framework, additional experiments in Section~\ref{subsub:fitness-function} analyze different values of 
$\alpha$ and show how the weighting factor can adjust the balance between attack effectiveness and stealthiness.

For crowd counting, semantic segmentation, and image fusion, the patch radius was set to 40, 40, and 30 pixels, with the number of colors in the patch color list set to 10, 10, and 2, respectively. The maximum number of iterations is uniformly set to 200 for all tasks.

\subsubsection{Datasets}
We conduct experiments on three publicly available datasets: the RGBT-CC dataset~\cite{liu2021cross}, the MF dataset~\cite{ha2017mfnet}, and the RoadScene dataset~\cite{tian2020road}. These datasets contain paired visible and infrared images and correspond to three representative tasks: crowd counting, semantic segmentation, and image fusion, respectively. This selection ensures both the diversity and representativeness of the experimental data.  
Following the protocol in reference~\cite{wei2023unified}, we randomly sample 100 images from the test set of each dataset using a fixed random seed with the selected samples covering diverse scenes to ensure reproducibility and cross-experiment comparability. 

\subsubsection{Target models}
We select several representative models from each task as the target models.~For the crowd counting task, BL+IADM~\cite{liu2021cross}, CAGNet~\cite{yang2024cagnet}, and CFAFNet~\cite{kong2024cross} were employed. For the semantic segmentation task, Openress~\cite{zhao2024open}, FEANet~\cite{deng2021feanet}, and SGFNet~\cite{wang2023sgfnet} were used. For the image fusion task, Res2Fusion~\cite{wang2022res2fusion}, UNFusion~\cite{wang2021unfusion}, and MaeFuse~\cite{li2025maefuse} served as the evaluation models.
All models are initialized and constructed using official pre-trained weights to ensure reliable and consistent experimental benchmarks.

\subsubsection{Evaluation metrics}

We evaluate attack methods from two perspectives: attack effectiveness and stealthiness. Details are as follows:

\textbf{Attack Effectiveness for Crowd Counting:} Grid Average Mean Absolute Error (GAME($k$))(Eq.~\ref{23})  and Root Mean Square Error (RMSE)(Eq.~\ref{24})~\cite{hodson2022root} are adopted as evaluation metrics: 
\begin{equation}
 \begin{aligned} 
 {\rm{GAME}}(\textit{k})=\frac{1}{n}\sum\limits_{i=1}^{n}{\sum\limits_{j=1}^{{{4}^{k}}}{|y_{\text{i}}^{j}-}\hat{y}_{i}^{j}|},
 \end{aligned}
 \label{23}
\end{equation}
\begin{equation}
 \begin{aligned} 
{\rm{RMSE}}=\sqrt{\frac{1}{n}\sum\limits_{i=1}^{n}{({{y}_{i}}-}{{{\hat{y}}}_{i}}{{)}^{2}}},
 \end{aligned}
 \label{24}
\end{equation}
where $n$ is the number of samples;  
$k$ determines the number of grids: $4^k$ indicates that the sample is divided into uniform $4^k$ sub-grids. $y_{\text{i}}^{j}$ and $\hat{y}_{i}^{j}$ denote the estimated counting and corresponding GT counts in region $j$ of image $i$, respectively; ${{y}_{i}}$ and ${{{\hat{y}}}_{i}}$ denote the estimated counting and corresponding GT counts of image $i$, respectively.
\textbf{Higher values} of GAME($k$) and RMSE indicate stronger degradation of the model’s prediction accuracy, \textbf{reflecting higher attack effectiveness}.

 \begin{table*}[!t]
\centering
\setlength{\tabcolsep}{3pt}
\caption{Comparison of experimental results with existing methods. Single-modal adversarial patch methods adapted to the cross-modal BL+IADM crowd counting model. “PSNR\_RGB” and “PSNR\_T” denote the PSNR values for visible images and infrared images, respectively. “SSIM\_RGB” and “SSIM\_T” denote the SSIM values for visible images and infrared images, respectively. \textbf{Bold} indicates the best results.}
\begin{tabular*}{\hsize}{@{\extracolsep{\fill}} c cccc c c cccc @{}}
\hline
\multicolumn{1}{l}{} & \multicolumn{5}{c}{Effectiveness} &  & \multicolumn{4}{c}{Stealthiness} \\ \cline{1-6} \cline{8-11} 
      & GAME(0)$\uparrow$ & GAME(1)$\uparrow$ & GAME(2)$\uparrow$ & GAME(3)$\uparrow$ & RMSE$\uparrow$   &  & PSNR\_RGB$\uparrow$ & SSIM\_RGB$\uparrow$ & PSNR\_T$\uparrow$ & SSIM\_T$\uparrow$ \\ \hline
Clean    & 13.7001 & 18.3601 & 22.1256 & 28.6380 & 24.4166 & & -                  & -                & -                 & -                 \\ 
PAP          & 14.9798 & 20.9912 & 26.6463 & 33.4813 & 25.1726 &  & 23.5981 & 0.9768 & 23.7474 & 0.9747 \\ 
APAM         & 14.6624 & 19.1693 & 23.1393 & 30.0921 & 25.9723 &  & \cellcolor{gray!20}\textbf{28.0505} & 0.9822 & 26.7175 & 0.9738 \\ 
AP-PCO(Ours) & \cellcolor{gray!20}\textbf{40.5543} & \cellcolor{gray!20}\textbf{51.2453} &\cellcolor{gray!20} \textbf{56.7172} &\cellcolor{gray!20} \textbf{63.6817} & \cellcolor{gray!20}\textbf{45.1786} & \textbf{} & 25.6450 &\cellcolor{gray!20} \textbf{0.9832} & \cellcolor{gray!20}\textbf{28.2151} & \cellcolor{gray!20}\textbf{0.9850} \\ \hline
\end{tabular*}
\label{table:9}
\end{table*}
\textbf{Attack Effectiveness for Semantic Segmentation:} Mean Intersection over Union (mIoU) (Eq.~\ref{25}) and Recall (Eq.~\ref{26}) are adopted as evaluation metrics: 
\begin{equation}
 \begin{aligned} 
{\rm{mIoU}}=\frac{1}{n-1}\sum\limits_{i=0}^{n}{({{P}_{ii}}/}(\sum\limits_{j=0}^{n}{({{p}_{ij}}+{{P}_{ji}})-{{P}_{ii}}))},
 \end{aligned}
 \label{25}
\end{equation}
\begin{equation}
 \begin{aligned} 
{\rm{Recall}}=\frac{1}{n+1}\sum\limits_{i=0}^{n}{({{P}_{ii}}/}\sum\limits_{j=0}^{n}{{{p}_{ji}})},
 \end{aligned}
 \label{26}
\end{equation}
where $n$ denotes the number of manually annotated object categories, and ${{P}_{ij}}$ is the number of pixels belonging to category $i$ that are predicted as category $j$. The combination of these two metrics provides a comprehensive assessment of the attack’s impact on model performance. Under adversarial conditions, \textbf{lower values} of mIoU and Recall \textbf{indicate stronger attack effectiveness}.

\textbf{Attack Effectiveness for Image Fusion:} Qabf~\cite{pang2023infrared}, SSIM, PSNR, Visual Information Fidelity (VIFF)~\cite{han2013new}, and Correlation Coefficient (CC)~\cite{liu2024infrared} are adopted as evaluation metrics. \textbf{Lower values} of Qabf, SSIM, PSNR, VIFF, and CC indicate stronger degradation of fusion quality, \textbf{reflecting higher attack effectiveness}.

\textbf{Attack Stealthiness:} To assess the stealthiness of the adversarial patches generated by the proposed method, two widely used image quality assessment metrics, SSIM and PSNR, are adopted. Higher PSNR values correspond to lower pixel-level distortion and greater visual similarity to the original image. SSIM ranges from 0 to 1, with values closer to 1 indicating stronger structural consistency between the adversarial and original images.

\subsection{Attacks in the Digital World}

\begin{table*}[!t]    
\centering
 \caption{Attack Effectiveness and Stealthiness of Our Method on Different Models for the Crowd Counting Task. “Clean” Denotes the Original Image, “Random” Denotes Images With Patches Added Using Randomly Sampled Positions and Colors.}
\resizebox{1\textwidth}{!}{
\begin{tabular*}{\hsize}{@{\extracolsep{\fill}} *{12}{@{}c@{}} @{}}
\hline
 \multirow{2}{*}{Model} & \multirow{2}{*}{Settings} & \multicolumn{5}{c}{Effectiveness} & & \multicolumn{4}{c}{Stealthiness} \\ 
\cline{3-7} \cline{9-12}
                            &                      & GAME(0)$\uparrow$ & GAME(1)$\uparrow$ & GAME(2)$\uparrow$ & GAME(3)$\uparrow$ & RMSE$\uparrow$ & & PSNR\_RGB$\uparrow$ & SSIM\_RGB$\uparrow$ & PSNR\_T$\uparrow$ & SSIM\_T$\uparrow$ \\ \hline 

\multirow{3}{*}{\makecell{BL+IADM}}    & Clean                & 13.70 & 18.36 & 22.13 & 28.64 & 24.42 & & -          & -        & -         & -        \\
                            & Random               & 14.20 & 18.80 & 22.80 & 29.32 & 24.36 & & \textbf{27.01} & \textbf{0.99} & \textbf{31.31} & \textbf{0.99} \\ 
                            & \cellcolor{gray!20} AP-PCO (Ours)  & \cellcolor{gray!20}\textbf{40.55} & \cellcolor{gray!20}\textbf{51.25} & \cellcolor{gray!20}\textbf{56.72} & \cellcolor{gray!20}\textbf{63.68} & \cellcolor{gray!20}\textbf{45.18} & \cellcolor{gray!20} & \cellcolor{gray!20}25.65 & \cellcolor{gray!20}0.98 & \cellcolor{gray!20}28.22 & \cellcolor{gray!20}0.99  \\  \cdashline{1-12}

\multirow{3}{*}{\makecell{CAGNet}}     & Clean                & 11.15 & 14.43 & 18.19 & 23.37 & 17.66 & & -          & -          & -         & -        \\
                            & Random               & 11.41 & 14.66 & 18.41 & 23.56 & 18.12 & & \textbf{27.04} & \textbf{0.99} & \textbf{31.28} & \textbf{0.99} \\
                            & \cellcolor{gray!20}AP-PCO (Ours)  & \cellcolor{gray!20}\textbf{21.05} & \cellcolor{gray!20}\textbf{25.78} & \cellcolor{gray!20}\textbf{29.69} & \cellcolor{gray!20}\textbf{34.31} & \cellcolor{gray!20}\textbf{29.20} & \cellcolor{gray!20} & \cellcolor{gray!20}24.77 & \cellcolor{gray!20}0.98 & \cellcolor{gray!20}26.84 & \cellcolor{gray!20}0.98   \\  \cdashline{1-12}

\multirow{3}{*}{\makecell{CFAFNet}}    & Clean                & 11.15 & 14.91 & 19.01 & 26.27 & 17.15 & & -          & -          & -         & -        \\
                            & Random               & 11.11 & 15.78 & 20.22 & 27.34 & 16.40 & & \textbf{26.73} & 0.95 & \textbf{31.11} & 0.96 \\       
                            & \cellcolor{gray!20}AP-PCO (Ours)  & \cellcolor{gray!20}\textbf{59.19} & \cellcolor{gray!20}\textbf{69.86} & \cellcolor{gray!20}\textbf{75.38} & \cellcolor{gray!20}\textbf{83.18} & \cellcolor{gray!20}\textbf{63.17} & \cellcolor{gray!20} & \cellcolor{gray!20}24.78 & \cellcolor{gray!20}\textbf{0.98} & \cellcolor{gray!20}26.77 & \cellcolor{gray!20}\textbf{0.98}   \\  \hline
\end{tabular*}
}
\label{table:1}
\end{table*}

\begin{figure*}[!t]
 	\centering
 	\includegraphics[width=1\textwidth]{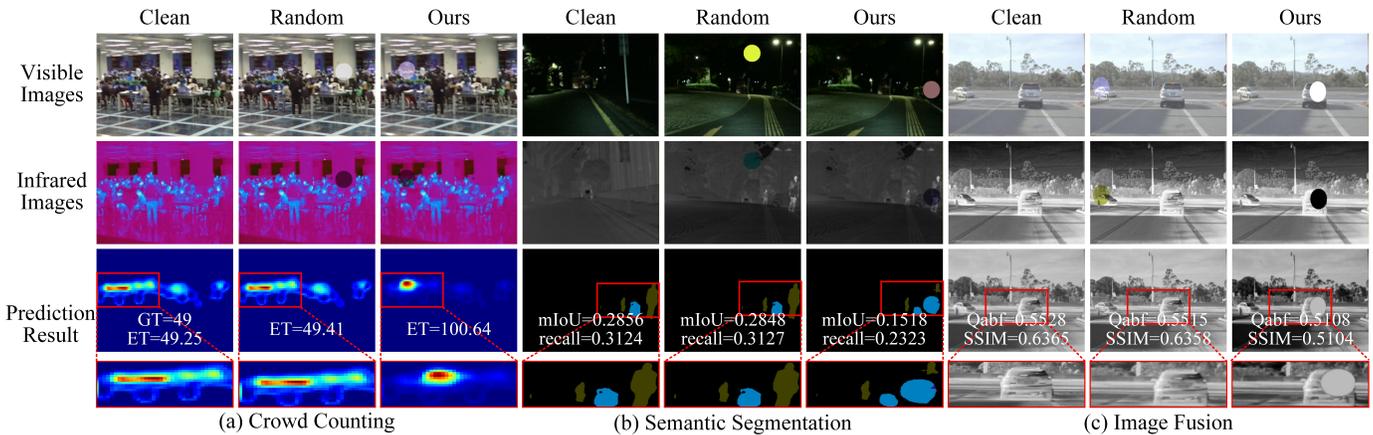}
 	\caption{Visualizations of patch attacks for the crowd counting, semantic segmentation, and image fusion tasks. As can be seen, the proposed method shows strong attack performance across three VI      dense prediction tasks.}
 	\label{keshihua}
 \end{figure*}
 \begin{figure*}[!t]
	\centering
	\includegraphics[width=1\textwidth]{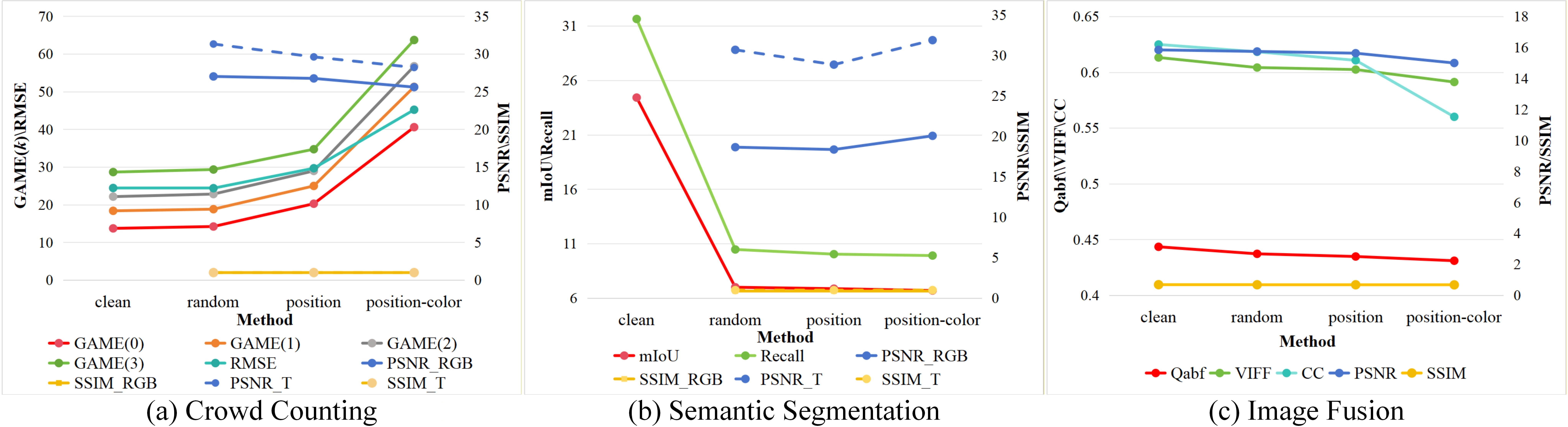}
	\caption{Comparison of ablation experiment results for the crowd counting task (BL+IADM model), the semantic segmentation task (Openress) and the image fusion task  (UNFusion). In this figure, “clean” denotes the original image; “random” denotes images with randomly added patches in both positions and color; “position” denotes images with patches optimized only for positions while color remains random; and “position-color” denotes images with patches optimized for both positions and color.}
	\label{Abl}
\end{figure*}
\begin{table}[!t] 
\centering
\small
 \caption{Attack effectiveness and Stealthiness of the Proposed Method on Different Models for the Semantic Segmentation Task.}
 \resizebox{1\columnwidth}{!}{ 
\begin{tabular}{ccccp{0.00001cm}cccc}

\hline
 \multirow{2}{*}{Model} & \multirow{2}{*}{Settings}      & \multicolumn{2}{c}{Effectiveness}  & \multicolumn{1}{c}{}& \multicolumn{4}{c}{Stealthiness}                                                        \\ \cline{3-4} \cline{6-9} 

                                 && mIoU$\downarrow$ 
                               & Recall$\downarrow$ &
                               & \makecell{PSNR\\\_RGB$\uparrow$}
                               &\makecell{SSIM\\\_RGB$\uparrow$} 
                               & \makecell{PSNR\\\_T$\uparrow$} 
                               & \makecell{SSIM\\\_T$\uparrow$} \\
                               \hline
\multirow{3}{*}{\makecell{Open\\ress}} & Clean  & 24.41 & 31.63  && - & - & - & - \\ 
                          & Random & 6.99  & 10.64 & & 18.65 & \textbf{0.86} & 30.69 & \textbf{0.98} \\ 
                          & \cellcolor{gray!20}AP-PCO (Ours)  & \cellcolor{gray!20}\textbf{6.69} & \cellcolor{gray!20}\textbf{9.90} & \cellcolor{gray!20} & 
                          \cellcolor{gray!20}\textbf{20.05} & \cellcolor{gray!20}0.86 & \cellcolor{gray!20}\textbf{31.88} & \cellcolor{gray!20}0.98 \\ \cdashline{1-9}

\multirow{3}{*}{\makecell{FEA\\Net}}   & Clean  & 52.52 & 70.68 &&  - & - & - & - \\ 
                          & Random & 50.32 & 68.06 & &\textbf{21.50} & 0.86 & \textbf{36.12} & \textbf{0.99} \\ 
                          & \cellcolor{gray!20}AP-PCO (Ours) & \cellcolor{gray!20}\textbf{1.58} & \cellcolor{gray!20}\textbf{10.91} & \cellcolor{gray!20} & 
                          \cellcolor{gray!20}20.45 & \cellcolor{gray!20}\textbf{0.86} & \cellcolor{gray!20}32.36 & \cellcolor{gray!20}0.98 \\ \cdashline{1-9}

\multirow{3}{*}{\makecell{SGF\\Net}}   & Clean  & 53.40 & 66.47 &&  - & - & - & - \\ 
                          & Random & 52.58 & 66.06 &  &\textbf{21.34} & \textbf{0.86} & \textbf{35.86} & \textbf{0.99} \\ 
                          & \cellcolor{gray!20}AP-PCO (Ours)  & \cellcolor{gray!20}\textbf{39.62} & \cellcolor{gray!20}\textbf{56.78} & \cellcolor{gray!20} &
                          \cellcolor{gray!20}20.72 & \cellcolor{gray!20}0.85 & \cellcolor{gray!20}33.35 & \cellcolor{gray!20}0.98 \\ \hline
\end{tabular}
}
\label{table:2}
\end{table}

\begin{table}[!t]
\small
\centering
 	\caption{Attack Effectiveness and Stealthiness of the Proposed Method on Different Models for the Image Fusion Task.}
 \resizebox{1\columnwidth}{!}{ 
\begin{tabular}{@{}lcccccc@{}}
\hline
                  Model                & Setting      & Qabf$\downarrow$                          & PSNR$\downarrow$                           & SSIM$\downarrow$                          & VIFF$\downarrow$                          & CC$\downarrow$      \\ \hline
\multirow{3}{*}{UNFusion}   & Clean & 0.52 & 13.89 & 0.65 & 0.74 & 0.61 \\ & Random & 0.51 & 13.88 & 0.64 & 0.72 & 0.60 \\
                                  & \cellcolor{gray!20}AP-PCO (Ours)    & \cellcolor{gray!20}\textbf{0.50} & \cellcolor{gray!20}\textbf{13.50} & \cellcolor{gray!20}\textbf{0.64} & \cellcolor{gray!20}\textbf{0.72} & \cellcolor{gray!20}\textbf{0.56} \\ \cdashline{1-7}

\multirow{3}{*}{Res2Fusion} & Clean & 0.44 & 15.84 & 0.68 & 0.61 & 0.62 \\ &Random        & 0.44 & 15.74 & 0.67 & 0.60 & 0.62 \\
                                  & \cellcolor{gray!20}AP-PCO (Ours)    & \cellcolor{gray!20}\textbf{0.43} & \cellcolor{gray!20}\textbf{15.00} & \cellcolor{gray!20}\textbf{0.67} & \cellcolor{gray!20}\textbf{0.59} & \cellcolor{gray!20}\textbf{0.56} \\\cdashline{1-7}
\multirow{3}{*}{MaeFuse}    & Clean & 0.46 & 15.81 & 0.68 & 0.56 & 0.65 \\ &Random        & 0.46 & 15.81 & 0.68 & 0.55 & 0.65 \\
                                  & \cellcolor{gray!20}AP-PCO (Ours)    & \cellcolor{gray!20}\textbf{0.46} & \cellcolor{gray!20}\textbf{15.66} & \cellcolor{gray!20}\textbf{0.67} & \cellcolor{gray!20}\textbf{0.55} & \cellcolor{gray!20}\textbf{0.64} \\ \hline

\end{tabular}
}
\label{table:3}
\end{table}

\subsubsection{Compare with other adversarial patch methods}
Existing adversarial patch attack methods for dense prediction tasks mainly focus on the visible modality, whose attack objectives, input forms, and optimization strategies are all based on the single-modality assumption. Consequently, they are difficult to be directly migrated to the VI dense prediction tasks with visible and infrared inputs, and extensive structural adjustments to the patch embedding method, optimization objective, and inter-modal feature alignment strategy are required when applied to cross-modal tasks. To verify their applicability, this paper selects two representative methods PAP \cite{liu2022harnessing} and APAM \cite{wu2021towards}, extends them to the cross-modal crowd counting BL+IADM model for comparative experiments without changing their core optimization mechanisms. As shown in Table ~\ref{table:9}, the experimental results indicate that single-modality patch methods fail to achieve stable and effective attack effects in VI dense prediction tasks.

\subsubsection{Attack performance against different models}
We evaluate the attack performance of the proposed method across multiple mainstream model architectures to assess its generalization and robustness. As shown in the experimental results in Tables~\ref{table:1}--\ref{table:3}, which correspond to the crowd counting, semantic segmentation, and image fusion tasks, and in the visualizations in Fig.~\ref{keshihua}, the optimized patches generated by our method achieve strong attack performance across all evaluated models. Although SSIM and PSNR values show a slight decrease compared with images containing randomly placed patches, this result is expected. Combined with the stronger attack effectiveness and visualization results, it indicates that the proposed method achieves a more reasonable trade-off between attack intensity and relative stealth, thereby verifying the effectiveness of the stealthiness enhancement strategy.

\begin{table}[!t]
\setlength{\tabcolsep}{1.5pt}
\renewcommand{\arraystretch}{1.1} 
\scriptsize
\caption{Stability analysis of adversarial attack performance for BL+IADM crowd counting model (mean ± std).}
\begin{tabular}{cccccc}
\hline
             & GAME(0)$\uparrow$ & GAME(1)$\uparrow$ & GAME(2)$\uparrow$ & GAME(3)$\uparrow$ & RMSE$\uparrow$ \\ \hline
clean        & 13.70$\pm$19.99                 & 18.36$\pm$19.90                 & 22.13$\pm$20.35                 & 28.64$\pm$22.21                 & 24.42$\pm$20.06              \\
AP-PCO(Ours) &\cellcolor{gray!20} \textbf{40.55$\pm$19.73}        & \cellcolor{gray!20}\textbf{51.25$\pm$19.83}        & \cellcolor{gray!20}\textbf{56.72$\pm$20.29}        & \cellcolor{gray!20}\textbf{63.68$\pm$21.13}        & \cellcolor{gray!20}\textbf{45.18$\pm$20.00}     \\ \hline
\end{tabular}
\label{table:mean}
\end{table}

\begin{table}[!t]
\centering
\setlength{\tabcolsep}{2pt} 
\renewcommand{\arraystretch}{1.1} 
\scriptsize
\caption{Effect of the Cross-modal Color Parameter Reuse Strategy on the Semantic Segmentation Task.}
\begin{tabular}{c c c c c c c c}
\hline
         & \multicolumn{2}{c}{Effectiveness} &  & \multicolumn{4}{c}{Stealthiness}       \\ \cline{2-3} \cline{5-8} 
         & mIoU$\downarrow$ 
                               & Recall$\downarrow$ &
                               & \makecell{PSNR\_RGB$\uparrow$}
                               &\makecell{SSIM\_RGB$\uparrow$} 
                               & \makecell{PSNR\_T$\uparrow$} 
                               & \makecell{SSIM\_T$\uparrow$}  \\ \hline
Not used & \cellcolor{gray!20}\textbf{36.06}           & \cellcolor{gray!20}\textbf{42.86}           &  & 20.4394  & 0.8543   & 26.2563 & 0.9802 \\
Used     & 39.62           & 56.78           &  & \cellcolor{gray!20}\textbf{20.7228}  & \cellcolor{gray!20}\textbf{0.8544}   & \cellcolor{gray!20}\textbf{33.3521} & \cellcolor{gray!20}\textbf{0.9835} \\ \hline
\end{tabular}
\label{table:cr}
\end{table}

\subsubsection{Ablation studies}To verify the effectiveness of each module in the proposed method, a series of ablation experiments is designed and conducted.

\textbf{Result Stability Analysis:} Although evaluation is conducted on a subset of 100 test images, the results exhibit stable and consistent degradation across samples. As shown in Table ~\ref{table:mean}, the standard deviation remains nearly unchanged after the attack. The result indicates consistent performance degradation across samples, meaning that the model’s prediction errors for all test samples increased by approximately the same margin at the original dispersion level, and there is no significant variation in attack effectiveness across samples. The relatively large variance mainly stems from the inherent density variation between sparse and highly congested scenes, which is common in crowd counting benchmarks.

\textbf{Effect of patch position-color optimization:} We conduct comparative experiments under two settings: one that applies the global search mechanism to position optimization only, and another that performs joint position and color optimization. As shown in Fig.~\ref{Abl}, the proposed joint optimization strategy produces adversarial patches with stronger attack effectiveness and also improves relative stealthiness in dense prediction tasks.

\textbf{Effect of cross-modal color parameter reuse:} We conduct comparative experiments with and without the color reuse mechanism. As shown in Table~\ref{table:cr}, the reuse strategy significantly improves stealthiness and avoids the noticeable artifacts that appear when visible-domain colors are directly applied to infrared images. The improvement is particularly evident in the infrared modality, where signal integrity is better preserved. Although attack effectiveness decreases slightly, this trade-off is expected because the reuse mechanism prioritizes stealthiness by suppressing excessive perturbations. Overall, the results demonstrate that the proposed strategy achieves a more balanced trade-off between attack strength and stealthiness.

\textbf{Effect of cross-modal attacks:}
We conduct ablation experiments using three strategies: attacking only the infrared modality, attacking only the visible modality, and attacking both modalities simultaneously. As shown in Table~\ref{table:8}, the cross-modal attack achieves stronger attack performance than either single-modality one, demonstrating the advantage of simultaneously perturbing both visible and infrared modalities.

\textbf{Effect of fitness function:}
\label{subsub:fitness-function}
We conduct comparative experiments on crowd counting and semantic segmentation tasks using two settings: one that considers only attack effectiveness (\(\alpha = 1\)) and another that jointly considers attack effectiveness and stealthiness (\(\alpha = 0.5\), \(\alpha = 0.1\)). 
Table~\ref{table:ac} and \ref{table:10} show the effectiveness of $\alpha$ in the fitness function for adjusting the trade-off between attack strength and stealthiness, and the framework can adapt to different task requirements.
\begin{table*} [!t] 
\centering
\setlength{\tabcolsep}{3.5pt}  
\footnotesize
\caption{Effect of Different Attack Modalities of Three Dense Prediction Tasks. “Visible” and “Infrared” Denote Attacks Applied to the Visible or Infrared Modality Only, While “VI” Denotes Attacks Applied to Both Modalities Simultaneously.}
 \resizebox{1\textwidth}{!}{ 
\begin{tabular*}{\hsize}{@{\extracolsep{\fill}} c *{5}{c} *{2}{c} *{5}{c} @{}}
\hline
\multirow{2}{*}{Modal} & \multicolumn{5}{c}{Crowd Counting} & \multicolumn{2}{c}{Semantic Segmentation} & \multicolumn{5}{c}{Image Fusion} \\ 
\cline{2-6} \cline{7-8} \cline{9-13}  
                       & \makecell{GAME(0)$\uparrow$} & \makecell{GAME(1)$\uparrow$} & \makecell{GAME(2)$\uparrow$} & \makecell{GAME(3)$\uparrow$} & \makecell{RMSE$\uparrow$} 
                       & \makecell{mIoU$\downarrow$} & \makecell{Recall$\downarrow$} 
                       & \makecell{Qabf$\downarrow$} & \makecell{PSNR$\downarrow$} & \makecell{SSIM$\downarrow$} & \makecell{VIFF$\downarrow$} & \makecell{CC$\downarrow$} \\ 
\hline
Visible    & 21.3622 & 25.8889 & 30.4938 & 36.2902 & 30.7622 & 6.97 & 9.96 & 0.4347 & 15.0853 & 0.6672 & 0.5956 & 0.5668 \\ 
Infrared     & 36.9179 & 47.8553 & 54.7607 & 62.1525 & 43.8502 & 7.56 & 10.75 & 0.4347 & 15.1014 & 0.6710 & 0.5963 & 0.5695 \\ 
VI 
& \cellcolor{gray!20}\textbf{40.5543} & \cellcolor{gray!20}\textbf{51.2453} & \cellcolor{gray!20}\textbf{56.7172} & \cellcolor{gray!20}\textbf{63.6817} & \cellcolor{gray!20}\textbf{45.1786} & \cellcolor{gray!20}\textbf{6.69} & \cellcolor{gray!20}\textbf{9.90} & \cellcolor{gray!20}\textbf{0.4309} & \cellcolor{gray!20}\textbf{14.9952} & \cellcolor{gray!20}\textbf{0.6659} & \cellcolor{gray!20}\textbf{0.5912} & \cellcolor{gray!20}\textbf{0.5600} \\ 
\hline
\end{tabular*}
}
\label{table:8}
\end{table*}

\begin{table*}[!htbp] 
\centering
\footnotesize
\caption{Effect of Different Fitness Function Weights on the Crowd Counting Task.}
\begin{tabular*}{\hsize}{@{\extracolsep{\fill}} c cccc cccccc @{}}
\hline
\multicolumn{1}{c}{} & \multicolumn{5}{c}{Effectiveness}                                                            &  & \multicolumn{4}{c}{Stealthiness}                                                        \\ \cline{1-6} \cline{8-11} 
$\alpha$ & GAME(0)$\uparrow$ & GAME(1)$\uparrow$ & GAME(2)$\uparrow$ & GAME(3)$\uparrow$ & RMSE$\uparrow$  &  & PSNR\_RGB$\uparrow$ & SSIM\_RGB$\uparrow$ & PSNR\_T$\uparrow$ & SSIM\_T$\uparrow$ \\ \hline
1    & \cellcolor{gray!20}\textbf{39.5844} & \cellcolor{gray!20}\textbf{53.0996} & \cellcolor{gray!20}\textbf{61.2997} & 68.8205 & 44.4267 & & 25.2850  & 0.9792  & 28.1450 & 0.9792 \\
0.5 & 39.2131         & 51.5339         & 61.2194         & \cellcolor{gray!20}\textbf{69.1666}         & \cellcolor{gray!20}\textbf{46.6531}   &      & 26.4535          & 0.9800         & 28.8502         & 0.9799              \\
0.1 & 30.8848         & 43.2991         & 50.7023         & 58.2660        & 38.2710   &      & \cellcolor{gray!20}\textbf{32.8543}         & \cellcolor{gray!20}\textbf{0.9855}         & \cellcolor{gray!20}\textbf{33.6389}        & \cellcolor{gray!20}\textbf{0.9837}             \\ \hline
\end{tabular*}
\label{table:ac}
\end{table*}

\begin{table}[!h] 
\centering
\setlength{\tabcolsep}{2pt} 
\renewcommand{\arraystretch}{1.1} 
\scriptsize
\caption{Experimental results on different fitness function weights for the semantic segmentation task. }
\begin{tabular}{ccclcccc}
\hline
\multicolumn{1}{l}{} & \multicolumn{2}{c}{Effectiveness}   &  & \multicolumn{4}{c}{Stealthiness}                                                        \\ \cline{1-3} \cline{5-8} 
    $\alpha$ & mIoU $\downarrow$                       & Recall$\downarrow$   &                    & PSNR\_RGB$\uparrow$                      & SSIM\_RGB$\uparrow$                     & PSNR\_T$\uparrow$                        & SSIM\_T$\uparrow$  \\ \hline
1   &\cellcolor{gray!20}\textbf{6.69} &\cellcolor{gray!20} \textbf{9.90}  & & 20.0540  & 0.8612   & 31.8827 & 0.9834 \\

0.5 &6.88      &10.36       &          &24.7938         &0.8734         &59.0620  &0.9996
\\
0.1 &6.88 &10.36  &  &\cellcolor{gray!20}\textbf{24.7985} &\cellcolor{gray!20}\textbf{0.8736} &\cellcolor{gray!20}\textbf{59.2910} &\cellcolor{gray!20}\textbf{0.9996}
\\ \hline
\end{tabular}
\label{table:10}
\end{table}

\begin{figure}
 	\centering
 	\includegraphics[width=1\columnwidth]{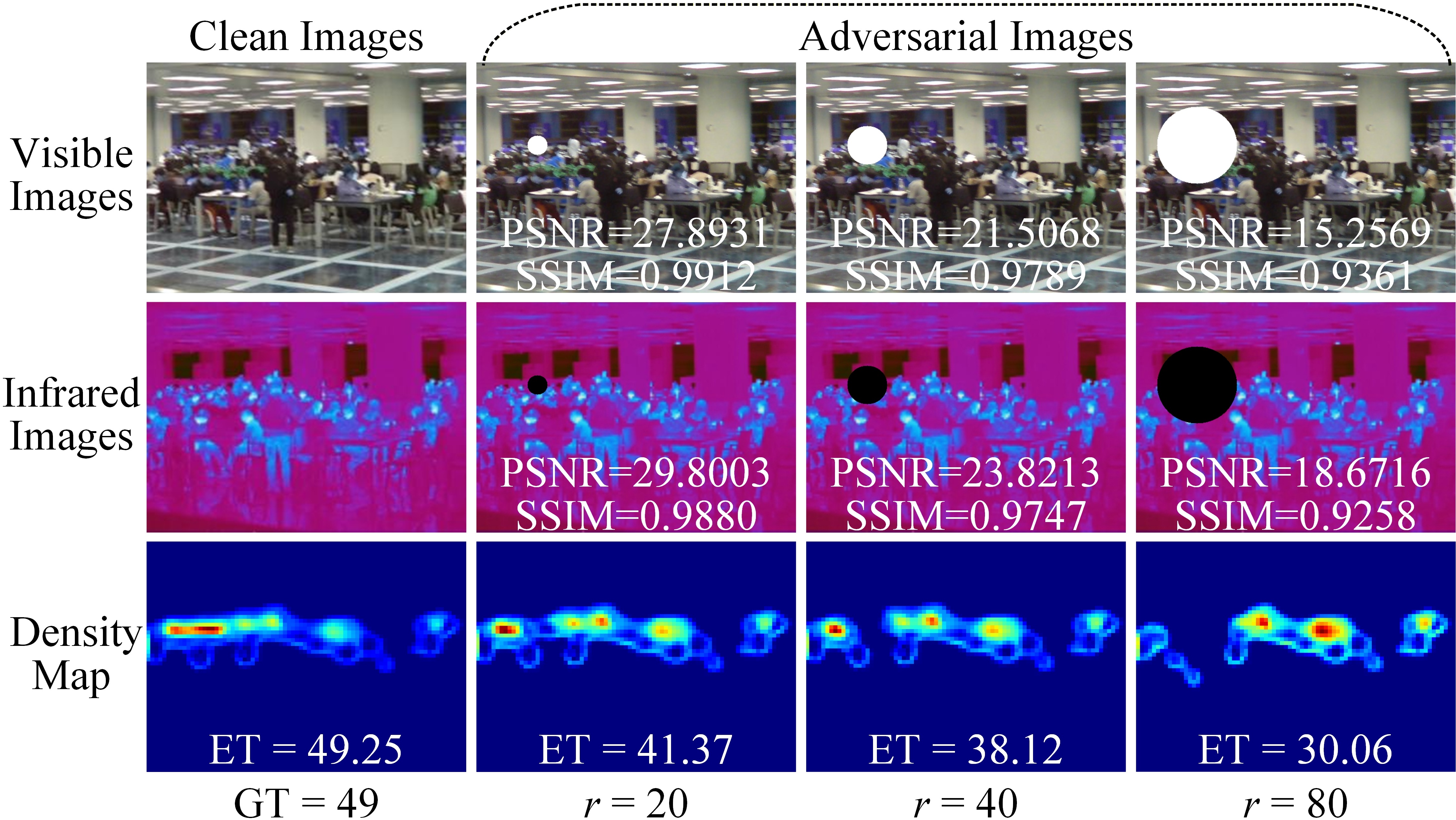}
 	\caption{Visualization of patch radius parameters attack performance. As the radius of the adversarial patch increases, its occlusion ratio in the image gradually rises, resulting in a significant degradation of stealthiness in cross-modal scenarios.}
 	\label{R}
 \end{figure}

\subsubsection{Hype-rparameter Settings}
We conduct the following experiments to investigate the impact of the patch hyper-parameters (iteration epochs, patch sizes, and the number of colors) on attack performance.

\begin{figure*}
 	\centering
 	\includegraphics[width=1\textwidth]{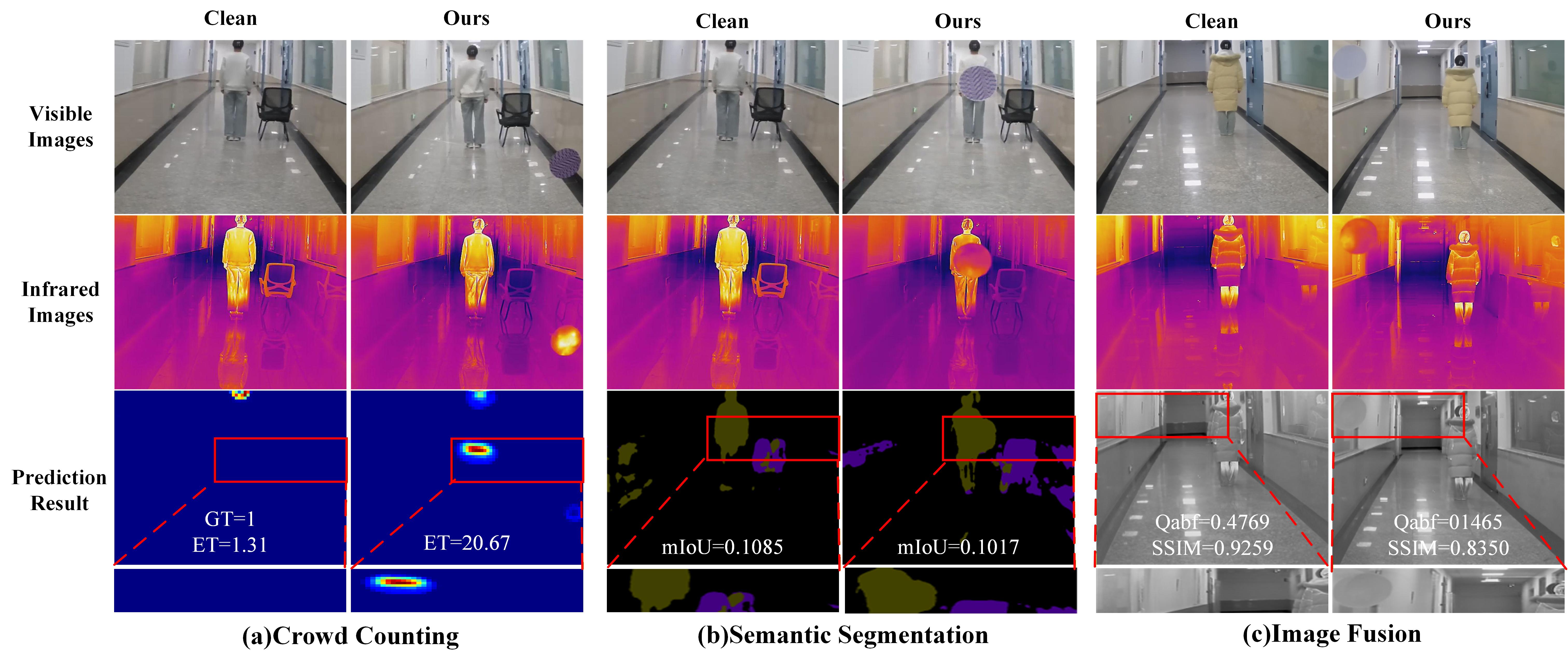}
 	\caption{Visualization results of adversarial patches against physical attacks in crowd counting, semantic segmentation, and image fusion tasks. As can be seen, the proposed method shows strong attack performance across three VI 
    dense prediction tasks.}
 	\label{wuli}
 \end{figure*}

\begin{table*}
\centering
\small
\setlength{\tabcolsep}{3pt}  
\caption{Experimental results on the patch radius parameter of the BL+IADM model for the crowd counting task. }
\begin{tabular*}{\hsize}{@{\extracolsep{\fill}} c cccc c c cccc @{}}
\hline
\multicolumn{1}{c}{} & \multicolumn{5}{c}{Effectiveness} &  & \multicolumn{4}{c}{Stealthiness} \\ \cline{1-6} \cline{8-11} 
      & GAME(0)$\uparrow$ & GAME(1)$\uparrow$ & GAME(2)$\uparrow$ & GAME(3)$\uparrow$ & RMSE$\uparrow$  &  & PSNR\_RGB$\uparrow$ & SSIM\_RGB$\uparrow$ & PSNR\_T$\uparrow$ & SSIM\_T$\uparrow$ \\ \hline
Clean & 13.7001 & 18.3601 & 22.1256 & 28.6380 & 24.4166 & & -          & -        & -         & -     \\
r=20  & 20.3511 & 25.2802 & 29.4354 & 35.0948 & 29.6377 & & \cellcolor{gray!20}\textbf{29.7634}   &\cellcolor{gray!20}\textbf{0.9922}    &\cellcolor{gray!20}\textbf{31.1696} &\cellcolor{gray!20}\textbf{0.9901}  \\
r=40  & 25.8050 & 30.3045 & 33.6848 & 38.7906 & 34.7603 & & 22.8577    & 0.9864   & 25.3934  & 0.9854   \\
r=80  & \cellcolor{gray!20}\textbf{34.1544} & \cellcolor{gray!20}\textbf{39.5820} &\cellcolor{gray!20}\textbf{42.4527} &\cellcolor{gray!20}\textbf{47.1011} & \cellcolor{gray!20}\textbf{44.3972} & & 17.4165   & 0.9518    & 20.2550 & 0.9475  \\ \hline
\end{tabular*}
\label{table:4}
\end{table*}

 \begin{table*}
\centering
\setlength{\tabcolsep}{3pt}  
\small  
\caption{Experimental results on the patch color number parameter of the BL+IADM model for the crowd counting task. }
\begin{tabular*}{\hsize}{@{\extracolsep{\fill}} c cccc c c cccc @{}}
\hline
\multicolumn{1}{l}{} & \multicolumn{5}{c}{Effectiveness} &  & \multicolumn{4}{c}{Stealthiness} \\ \cline{1-6} \cline{8-11} 
      & GAME(0)$\uparrow$ & GAME(1)$\uparrow$ & GAME(2)$\uparrow$ & GAME(3)$\uparrow$ & RMSE$\uparrow$   &  & PSNR\_RGB$\uparrow$ & SSIM\_RGB$\uparrow$ & PSNR\_T$\uparrow$ & SSIM\_T$\uparrow$ \\ \hline
Clean    & 13.7001 & 18.3601 & 22.1256 & 28.6380 & 24.4166 & & -                  & -                & -                 & -                 \\
Color=1 & 25.8050 & 30.3045 & 33.6848 & 38.7906 & 34.7603 & & 22.8577    &\cellcolor{gray!20}\textbf{0.9864}     &25.3934  &\cellcolor{gray!20}\textbf{0.9854}      \\
Color=2  & 25.2305 & 30.2875 & 33.9155 & 39.0918 & 33.5994 & & 23.8335          & 0.9816          & 25.8799          & 0.9797          \\
Color=5  & 28.8294 & 34.7558 & 39.1652 & 45.1211 & 36.9115 & & 24.4387          & 0.9836 & 27.1367          & 0.9848          \\
Color=10 &\cellcolor{gray!20} \textbf{39.5943} & \cellcolor{gray!20}\textbf{51.1300} & \cellcolor{gray!20}\textbf{56.8469} & \cellcolor{gray!20}\textbf{63.9860} & \cellcolor{gray!20}\textbf{44.1738} & & \cellcolor{gray!20}\textbf{25.6450} & 0.9832          & \cellcolor{gray!20}\textbf{28.2151} & 0.9850 \\ \hline
\end{tabular*}
\label{table:5}
\end{table*}

\begin{table}
\centering
\small
\setlength{\tabcolsep}{2pt} 
\renewcommand{\arraystretch}{1.1} 
\scriptsize
\caption{Experimental results on the patch color number parameter of the openress model for the semantic segmentation task.}
\begin{tabular}{ccclcccc}
\hline
\multicolumn{1}{l}{} & \multicolumn{2}{c}{Effectiveness}   &  & \multicolumn{4}{c}{Stealthiness}                                                        \\ \cline{1-3} \cline{5-8} 
         & mIoU $\downarrow$                       & Recall$\downarrow$    &                   & PSNR\_RGB$\uparrow$                      & SSIM\_RGB$\uparrow$                     & PSNR\_T$\uparrow$                        & SSIM\_T$\uparrow$                       \\ \hline
Clean   & 24.41         & 31.63     &    & -                                  & -                 & -                 & -\textbf{}                \\
Color=1 & \cellcolor{gray!20}\textbf{6.65} & 9.96  &        & 19.1507                          & \cellcolor{gray!20}\textbf{0.8645} & 29.8579          & 0.9833          \\ 
Color=2                         & 6.71          & 9.94 &         & 19.3025                          & 0.8625          & 30.3935          & 0.9829          \\ 
Color=5                         & 6.69          & 9.92   &       & 19.7081                          & 0.8614 & 30.8039          & \cellcolor{gray!20}\textbf{0.9835} \\
Color=10                        & 6.69 & \cellcolor{gray!20}\textbf{9.90} & & \cellcolor{gray!20}\textbf{20.0540}                 & 0.8612          & \cellcolor{gray!20}\textbf{31.8827} & 0.9834 \\ \hline
 \label{table:6}
\end{tabular}
\end{table}

\begin{table}  
\centering
\small
\caption{Experimental results on the patch color number parameter of the Res2Fusion model for the image fusion task.}
\begin{tabular*}{\linewidth}{@{\extracolsep{\fill}} l cccc c @{}}
\hline
         & Qabf$\downarrow$                          & PSNR$\downarrow$                           & SSIM$\downarrow$                          & VIFF$\downarrow$                          & CC$\downarrow$                            \\  \hline
Clean   & 0.4434          & 15.8424          & 0.6784                      & 0.6132          & 0.6249          \\ 
Color=1 & \cellcolor{gray!20}\textbf{0.4305} & 15.0364          & 0.6662                      & \cellcolor{gray!20}\textbf{0.5912} & 0.5618          \\ 
Color=2 & 0.4309          & \cellcolor{gray!20}\textbf{14.9952} & 0.6659                      &\cellcolor{gray!20} \textbf{0.5912} &\cellcolor{gray!20} \textbf{0.5600} \\ 
Color=5 & 0.4393          & 15.2879          &\cellcolor{gray!20} \textbf{0.6613}             & 0.6005          & 0.5847          \\
Color=10& 0.4393          & 15.2915          & 0.6615                      & 0.6002          & 0.5843          \\ \hline
\end{tabular*}
\label{table:7}
\end{table}

\textbf{Patch Sizes}:
We control the patch size by adjusting the radius $r$ of the circular patch to 20, 40, and 80. As shown in Table~\ref{table:4}, both GAME and RMSE increase as the patch size grows, indicating enhanced attack effectiveness. However, the occlusion rate also increases, reducing stealthiness. As shown in Fig.~\ref{R}, larger patches cover more irrelevant regions of the image and are more perceptible. Considering the trade-off between attack effectiveness and stealthiness, we choose a radius of $r=40$. For the semantic segmentation and image fusion tasks, the patch radius is set to 40 and 30 pixels based on the respective image sizes.

\textbf{Color Numbers}: We conduct experiments using 1, 2, 5, and 10 colors. Based on the results in Tables~\ref{table:5}, \ref{table:6}, and \ref{table:7}, we select 10 colors for the crowd counting and semantic segmentation tasks, and 2 colors for the image fusion task.

\textbf{Iteration Epochs}: The epochs represent the maximum number of evolution iterations for a single image. Although increasing epochs generally improves attack performance via broader search, it drastically raises computational cost. Therefore, the number of epochs is first set to 200, and an early stopping mechanism is adopted to automatically stop optimization for a sample when its mean fitness value remains unchanged for 10 consecutive generations (the fitness function converges). Experimental results show that approximately 12\%, 60\%, and 21\% of samples converged early on the crowd counting, semantic segmentation, and image fusion tasks, respectively. This indicates that some samples can achieve stable attack performance with fewer iterations, while setting the maximum number of iterations to 200 can provide sufficient search space for the remaining samples to enhance attack performance. Considering both attack performance and time efficiency, the number of epochs is set to 200.

\subsection{Attacks in the Physical World}
To verify the effectiveness of the adversarial patches in the physical world, we select a laboratory corridor as a representative scenario to conduct real-world image capture and physical attack validation, as shown in Fig.~\ref{wuli}. Experimental results demonstrate that the generated adversarial patches exhibit attack performance in the three tasks: crowd counting, semantic segmentation, and image fusion. If applied to practical scenarios, they will pose potential security risks to the relevant fields relying on VI 
dense prediction systems.

 \begin{table}
\centering
\setlength{\tabcolsep}{3pt}
\footnotesize
\caption{Experimental Results of Defense Methods on the Crowd Counting Task.}
\begin{tabular}{cccccc}
\hline
       & GAME(0) & GAME(1) & GAME(2) & GAME(3) & RMSE    \\ \hline
clean  & 13.7001 & 18.3601 & 22.1256 & 28.6380 & 24.4166 \\

JPEG   & 30.0835 & 40.1116 & 45.5558 & 52.6953 & 36.6276 \\
MF     & 30.2246 & 41.7568 & 47.1867 & 54.3790 & 36.6903 \\AP-PCO(Ours) & \cellcolor{gray!20}\textbf{40.5543} & \cellcolor{gray!20}\textbf{51.2453} & \cellcolor{gray!20}\textbf{56.7172} &\cellcolor{gray!20} \textbf{63.6817} &\cellcolor{gray!20} \textbf{45.1786} \\ \hline
\end{tabular}
\label{table12}
\end{table}

\subsection{Defense Against Adversarial Patches}
This paper verifies three typical defense methods: Joint Photographic Experts Group (JPEG) compression\cite{liu2019feature}, median filter(MF)\cite{gu2019detecting}, and mean square error (MSE)-based anomaly detection\cite{li2025generalizable}. The experimental results are shown in Table ~\ref{table12}, and none of the methods achieve the defense effect: JPEG compression removes high-frequency image components and disrupts the fine textures and high-frequency perturbation features of adversarial patches. MF smooths local pixels and suppresses sharp noise patterns, slightly weakening the attack performance. The MSE-based anomaly detection method identifies attack samples by calculating the mean square error between the predicted density map and the real annotation, and can only detect 22\% of the attack samples. In summary, the experiments demonstrate that the adversarial patches proposed in this paper have strong robustness and can effectively resist a variety of classical defense strategies.

\section{Conclusions}
This work presents a cross-modal joint optimization framework that uses global search to optimize the positions and colors of adversarial patches for visible and infrared images. A cross-modal color reuse strategy adapts a shared set of color parameters to both modalities, enabling effective disruption of multimodal feature representations while maintaining low saliency. Experiments on crowd counting, semantic segmentation, and image fusion show that the generated patches consistently degrade performance across diverse network architectures and effectively evade representative existing defense methods. These results verify the effectiveness and generalizability of cross-modal adversarial patch attacks and their prominent robustness further reveals the security vulnerabilities inherent in existing multimodal perception systems. Overall, this study advances the understanding of multimodal AI security and provides a practical tool for evaluating the vulnerability of multimodal perception models. It should be noted that our method still has certain limitations: physical world interference factors, such as variations in camera viewing angles and illumination conditions, may degrade the attack effectiveness of cross-modal adversarial patches in real-world scenarios. A systematic investigation and validation of these physical influencing factors will serve as an important direction for future work.

\bibliography{main}
\bibliographystyle{IEEEtran}

\vfill

\end{document}